\documentclass[letterpaper, 10 pt, conference]{ieeeconf}
\usepackage{times}
\usepackage[pdftex]{graphicx}
\usepackage{subfigure}
\usepackage{amsmath,amssymb,amsopn,amstext,amsfonts}
\usepackage{cancel}
\usepackage[space]{cite}
\usepackage{pdfsync}
\usepackage{balance}
\usepackage{color}
\usepackage{mathtools}
\usepackage{multirow}
\usepackage{makecell}
\usepackage{algorithm}  
\usepackage{algorithmicx}  
\usepackage{algpseudocode} 
\usepackage{bm}

\usepackage{diagbox}
\usepackage{float}
\usepackage{epstopdf}
\usepackage{pifont}
\usepackage{multirow}
\usepackage{url}
\usepackage{tabularx}
\usepackage[linkcolor=black,citecolor=black,urlcolor=black,colorlinks=true]{hyperref}
\usepackage{verbatim} 
\usepackage[skip=3pt, font={small}]{caption}

\graphicspath{{../figure/}}
\DeclareGraphicsExtensions{.png,.jpg,.eps,.pdf}
\IEEEoverridecommandlockouts
\title{\LARGE \bf QuadHand: A CoG-Compensated Quadrotor Aerial Manipulator \\ with Whole-Body Motion Planning
}

\title{\LARGE \bf QuadHand: A Compact Quadrotor Aerial Manipulator with MRC-SDF-Based Whole-Body Motion Planning}

\author{ Rui Jin$^{*}$, Ruiyang Liu$^{*}$, Xinhang Xu, Haotian Jin, Yi Wang, Yizhuo Yang, and Lihua Xie$^{\dagger}$%
    \thanks{ $^*$Indicates equal contribution. }
	\thanks{$^\dagger$Corresponding author: {\tt\small elhxie@ntu.edu.sg}.}
	\thanks{
	Rui Jin, Ruiyang Liu, Xinhang Xu, Haotian Jin, Yi Wang, and Yizhuo Yang are with the School of Electrical and Electronic Engineering, Nanyang Technological University, Singapore 639798.
	Lihua Xie is with the NTU--VinUni Joint Research Laboratory for Embodied AI and Robotics, School of Electrical and Electronic Engineering, Nanyang Technological University, Singapore 639798, and VinUniversity, Hanoi, Vietnam.
	}
    \thanks{
    This work was supported by Ministry of Education, Singapore, under AcRF TIER 1 Grant RG64/23.
	} 
}

\makeatletter
\let\@oldmaketitle\@maketitle

\makeatother

\begin{document}

\maketitle
\thispagestyle{empty}
\pagestyle{empty}

\begin{abstract}
Uncrewed aerial manipulators (UAMs) integrate robotic arms with aerial platforms for three-dimensional physical interaction.
However, enlarging the workspace increases arm-induced disturbances, while existing geometric representations face a trade-off between geometric fidelity and computational efficiency in close-proximity interaction.
This paper presents QuadHand, a compact quadrotor aerial manipulator with a 3-DoF arm, gripper, and battery-assisted passive CoG compensation module to reduce dominant arm-induced disturbances.
We further propose MRC-SDF, a Multi-articulated Robot-Centric Signed Distance Field that preserves fine geometric detail with tractable computation, and a spatiotemporal whole-body trajectory optimization framework that jointly optimizes the quadrotor and manipulator for safe and executable trajectory generation.
Simulations and real-world experiments demonstrate safe and executable aerial manipulation in complex environments.
\end{abstract}

\section{Introduction}
\label{sec:intro}
Uncrewed Aerial Manipulators (UAMs) integrate robotic arms with aerial platforms, transforming drones from observation-centric platforms~\cite{zhang2024npe, zhang2025grounded, yao2026arsgaussian, jin2024gs} into systems capable of physical interaction in three-dimensional mid-air workspaces.
They have been applied to grasping~\cite{wu2026hand}, maintenance~\cite{jinjie25teleoperate}, transportation~\cite{jin2025tethered, sun2025agile, li2026aerothrow}, and assembly~\cite{cao2024aircrab}.
However, close-range interaction remains difficult to teleoperate due to the high DoFs of the coupled base--arm system, requiring coordinated base--arm motion for safe and precise interaction.
We identify three key challenges.

The first challenge is how to achieve a larger end-effector workspace. 
UAMs require the manipulator to provide as large a reachable workspace as possible to accomplish a wider range of tasks. However, as the workspace increases, both the interaction forces at the end-effector and the manipulator's own motions can induce larger disturbances on the aerial base. Achieving a favorable balance between workspace and base disturbance is therefore crucial for improving end-effector control accuracy in UAMs.

\begin{figure}[t]
    \begin{center}
        \includegraphics[width=1.0\columnwidth]{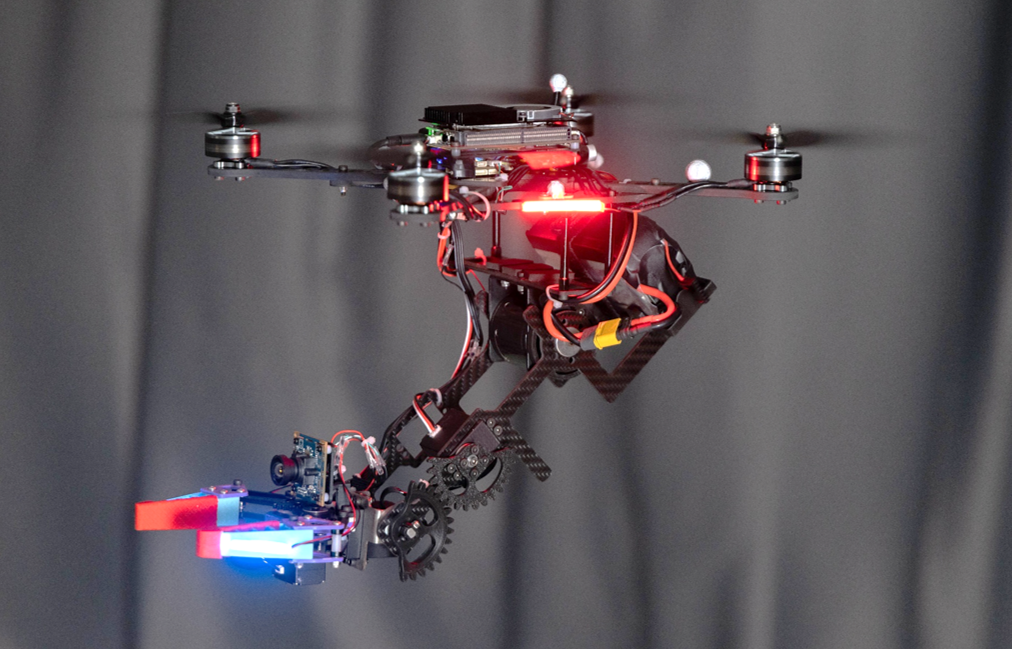}
    \end{center}
    \caption{The proposed QuadHand system hovering in mid-air.}
    \vspace{-0.5 cm}
    \label{fig:head}
\end{figure}

The second challenge is minimizing information loss in geometric representation. Unlike navigation-only drones, UAMs operate in close proximity to objects for interaction, which significantly constrains the feasible solution space. Therefore, designing appropriate geometric representations of both the robot and the environment is essential to ensure safe and precise operation in complex environments.

The third challenge is how to plan in high-dimensional configuration spaces. In UAM systems, the mounting of a robotic arm onto an aerial platform results in a system with high DoFs. In such a configuration, the aerial base and the manipulator are kinematically and dynamically coupled. Consequently, simply decomposing this high-dimensional configuration space and treating the base and arm as independent systems restricts the available solution space, often resulting in suboptimal or infeasible trajectories.

Based on the aforementioned observations and analyses, we propose a comprehensive aerial manipulation system that integrates a quadrotor platform with a 3-DoF robotic arm and a gripper, together with a whole-body trajectory optimization framework and a high-fidelity geometric representation method, as shown in Fig.~\ref{fig:head}.
Specifically, to maintain an essential operational workspace while alleviating base disturbances, QuadHand integrates a quadrotor equipped with a 3-DoF manipulator and gripper, and adopts a battery-assisted Passive Center-of-Gravity Compensation (PCGC) module, inspired by prior CoG-compensation mechanisms~\cite{trujillo2019novel,suarez2021cartesian,ollero2021past}, to help mitigate the dominant CoG shift induced by arm motion.
This design allows the end-effector to extend outward effectively while mitigating the increase in the moment arm.
To balance geometric fidelity and computational efficiency for whole-body trajectory optimization with close-proximity interaction in cluttered environments, we introduce a Multi-articulated Robot-Centric Signed Distance Field (MRC-SDF) representation.
By using pre-generated self signed distance fields (SDFs) for the base and each manipulator link and preserving the original environmental point cloud, the system retains rich geometric detail without the overhead of real-time flight-corridor or ESDF generation, thereby facilitating safe and executable whole-body trajectories in extreme proximity to target objects or environmental features. 
Furthermore, to enhance motion planning in high-dimensional configuration spaces, we propose an efficient spatiotemporal whole-body trajectory optimization framework that performs coupled base-arm optimization to generate safe and executable trajectories for diverse manipulation tasks. 
To validate the proposed system, we conducted extensive simulations and real-world experiments, and the results demonstrate its effectiveness. The main contributions of this paper are summarized as follows:
\begin{enumerate}
    \item A compact QuadHand aerial manipulation platform that integrates a quadrotor, a 3-DoF arm, a gripper, and a battery-assisted passive CoG compensation module for validating whole-body aerial manipulation.
    \item The MRC-SDF geometry representation method minimizes geometric information loss, enabling close-proximity interaction trajectory generation.
    \item An efficient spatiotemporal whole-body trajectory optimization framework for coupled base--arm planning.
    \item Extensive simulations and real-world flight tests demonstrating the system's efficiency.
\end{enumerate}

\begin{figure}[!t]
	\begin{center}
		\includegraphics[width=1.0\columnwidth]{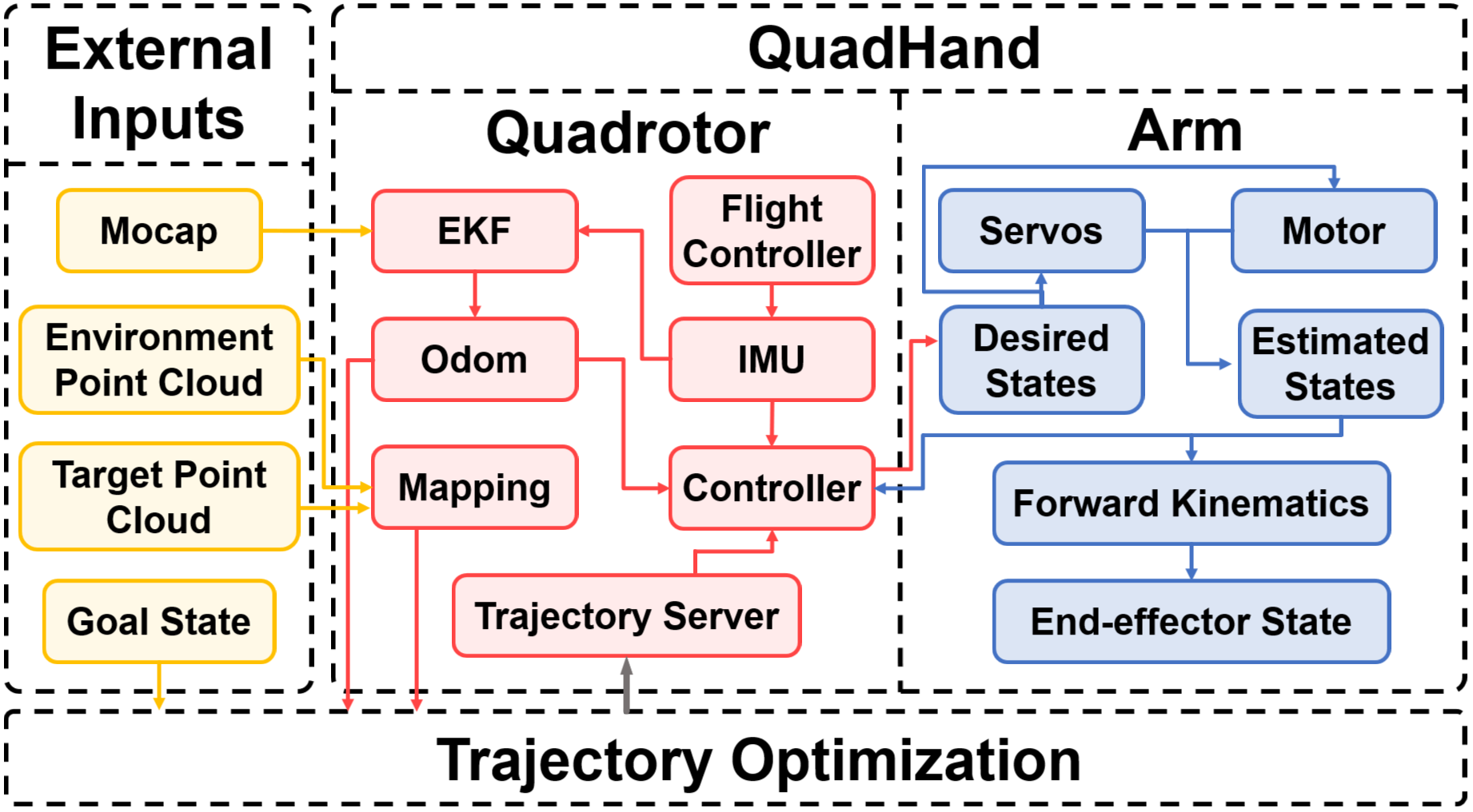}
	\end{center}
	\caption{
		\label{fig2} An overview of our proposed aerial manipulation system.}
	\vspace{-1.2cm}
\end{figure}

\begin{figure*}[!t]
    \centering
    \includegraphics[width=2.0\columnwidth]{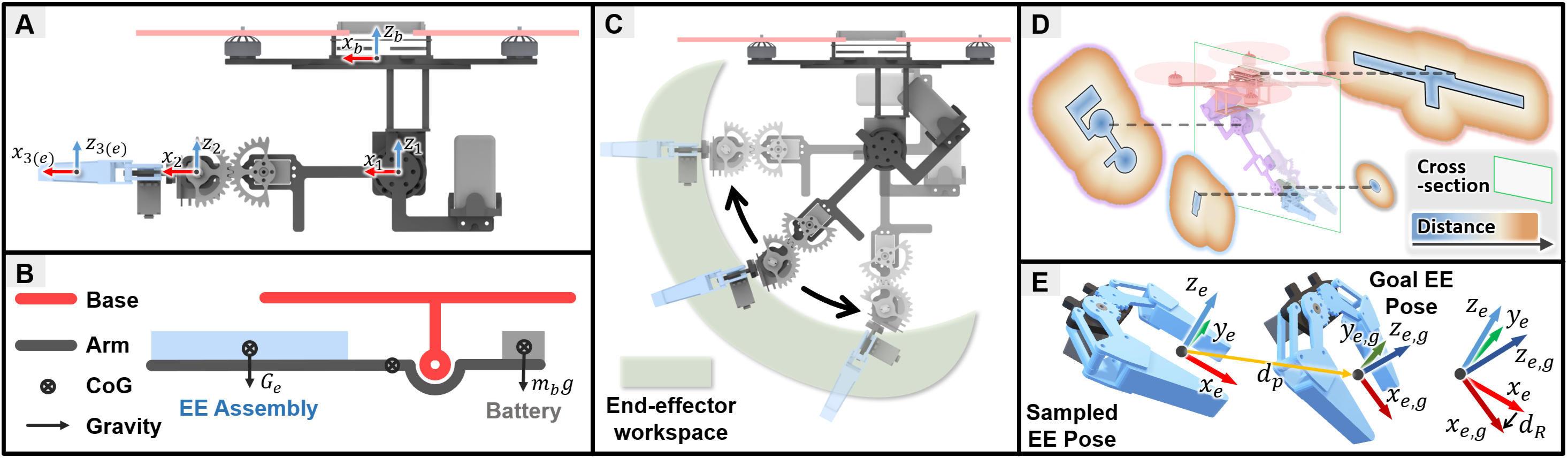}
    \caption{Overview of the UAM platform and the key components in the proposed planning framework. (A) Coordinate definitions. 
    (B) PCGC illustration: as the arm moves, the battery passively rotates, synchronously adjusting the gravity moment arm. (C) End-effector workspace. (D) MRC-SDF illustration, where each link is represented by a precomputed SDF. (E) Penalty term for end-effector insertion: the sampled end-effector poses along the trajectory are encouraged to match a predefined pose.}
    \label{fig:3}
    \vspace{-0.5cm}
\end{figure*}

\section{Related works}
\label{sec:related_works}
\subsection{Quadrotor Aerial Manipulation Platform}
\label{subsec:am_platform}

Aerial manipulation platforms combine flying bases, such as quadrotors~\cite{cao2024aircrab, hongming25ICRA, cao2025nature, meng25tro, sihao25quaduam, zhaopeng25tmech}, hexarotors~\cite{he2025flying, gupta2025umi, khj25ijrr}, or ducted fans~\cite{junxiao25iros}, with manipulators.
Quadrotors are widely used due to mature hardware, compactness, and ease of integration, and commonly carry serial~\cite{meng25tro,zhaopeng25tmech,sihao25quaduam} or parallel~\cite{hongming25ICRA,cao2025nature} manipulators.
Serial arms provide larger workspace and dexterity but introduce longer moment arms, higher joint torques, and larger CoG shifts, whereas parallel manipulators offer higher positioning accuracy~\cite{tsai1999robot} but limited workspace.
Related quadrotor studies have further addressed configuration-dependent robust control and actuator-constrained geometric control~\cite{derrouaoui2025improved,huang2025geometric}.
These trade-offs motivate compact quadrotor manipulators that retain workspace while reducing arm-induced CoG disturbances.
Following CoG-compensation studies~\cite{trujillo2019novel,suarez2021cartesian,ollero2021past}, QuadHand integrates a battery-assisted passive CoG compensation module.

\subsection{Geometric Representations for Whole-Body Planning}
\label{subsec:geo_planning}
Aerial manipulation planning relies on appropriate geometric representations of both the aerial manipulator and the surrounding environment to enable close-range interactions with target objects.

Many works represent the environment using corridors to enable whole-body planning. For example, Deng et al.\cite{deng25tro} approximated the aerial platform as an ellipsoid and modeled the environment with flight corridors, while Zhang et al.\cite{zhaopeng25tmech} represented the robot as a dynamic convex hull and modeled the environment with convex regions. While effective for navigation, these abstractions inevitably introduce geometric loss on either the robot body or the target object, making it difficult to explicitly reason about close-range interactions between the end-effector and the object, where contact clearance and local geometry become critical.

ESDFs are also used to model the environment while the robot is approximated by sampled points, enabling whole-body trajectory optimization~\cite{mengke23icra, chengkai24icra}.
With enough resolution, ESDF preserves geometric details. However, fine manipulation typically requires a higher-resolution ESDF, increasing memory cost and slowing construction.

\section{System Architecture}
\label{sec:system_architecture}
Fig.~\ref{fig2} illustrates the overall architecture of the proposed QuadHand, which consists of a quadrotor base and a 3-DoF robotic arm (Sec.~\ref{sec:hardware_design}). 
The quadrotor state is estimated by an Extended Kalman Filter (EKF) that fuses pose measurements from a motion capture system with onboard IMU data. 
The robotic arm executes joint-level control commands and provides real-time state feedback. 
The UAM is modeled as MRC-SDF (Sec.~\ref{sec:sdf}) and incorporated into trajectory optimization to generate whole-body trajectories with collision avoidance and kinematic feasibility (Sec.~\ref{sec:traj_opt}).

\section{Hardware Design}
\label{sec:hardware_design}

To mitigate the dominant CoG shift introduced by arm motion, we integrate a battery-assisted passive compensation module, inspired by prior CoM/CoG compensation designs~\cite{trujillo2019novel,suarez2021cartesian,ollero2021past}, into a compact quadrotor-based aerial manipulator with a 3-DoF arm and gripper, as shown in Fig.~\ref{fig:3}.A-C.

The proposed QuadHand weighs approximately 1.7~kg and consists of two main subsystems. 
The upper subsystem is a quadrotor with a 330~mm wheelbase, carrying the onboard computer, flight controller, ESC, and related electronics, and providing the lift and control torques required for aerial motion.
The lower subsystem comprises a 3-DoF robotic arm with two pitch joints and one roll joint, and a gripper mounted at the arm’s end for grasping and interaction tasks.

During motion of the first joint, the battery passively follows the manipulator movement, adjusting the platform’s CoG to stay near the wheelbase center, thereby reducing disturbances to the aerial base.

\section{Multi-Articulated Robot-centric SDF}
\label{sec:sdf}
As analyzed in Sec.~\ref{sec:intro}, when UAMs navigate cluttered environments and interact with target objects, they face an inherent trade-off between geometric accuracy and computational efficiency. Inspired by~\cite{shuang2023iros}, we introduce the MRC-SDF representation, which enables direct distance queries between the UAM’s whole-body configuration and surrounding obstacles via preconstructed SDFs while preserving the environment’s original point-cloud geometry.

In the proposed MRC-SDF, each collision-checking link of the UAM is associated with a precomputed SDF, denoted as $\mathrm{SDF}_i(\cdot)$ for $i=0,\ldots,N_l-1$, where $N_l$ denotes the number of link-wise SDFs used for collision checking, as illustrated in Fig.~\ref{fig:3}.D.
For the $j$-th obstacle point $\mathbf{x}_j \in \mathbb{R}^3$ expressed in the world frame, we first express it in the $i$-th link frame as ${}^{i}\mathbf{x}_j$:
\begin{equation}
\begin{bmatrix}
{}^{i}\mathbf{x}_j\\
1
\end{bmatrix}
=
\left({}^{w}\mathbf{T}_{i}\right)^{-1}
\begin{bmatrix}
\mathbf{x}_j\\
1
\end{bmatrix},
\label{eq:point_transform_T}
\end{equation}
where ${}^{w}\mathbf{T}_{i}=\begin{bmatrix}{}^{w}\mathbf{R}_{i} & {}^{w}\mathbf{t}_{i}\\ \mathbf{0}^{\top} & 1\end{bmatrix}\in SE(3)$ is the pose of link $i$ in the world frame. The signed distance from $\mathbf{x}_j$ to link $i$ is then obtained by
\begin{equation}
{}^{i}d_j = \mathrm{SDF}_i\!\left({}^{i}\mathbf{x}_j\right).
\label{eq:link_sdf}
\end{equation}

\section{Trajectory Optimization}
\label{sec:traj_opt}
As introduced in Sec. \ref{sec:intro}, the UAM requires efficient trajectory planning in a high-dimensional state space. 
This section aims to formulate and solve the whole-body trajectory generation problem while considering dynamic feasibility, ensuring safety from free flight to task completion.

\subsection{Dynamic Model and Differential Flatness}

In this paper, we adopt a simplified UAM model, whose base state consists of the translation $\mathbf{p}_b = (p_x, p_y, p_z)^{\top} \in \mathbb{R}^3$ and the rotation $\mathbf{R}_b \in SO(3)$, while the manipulator state includes the joint angles $\boldsymbol{\theta} \in \mathbb{R}^{N_{\theta}}$ and the corresponding joint angular velocities $\dot{\boldsymbol{\theta}} \in \mathbb{R}^{N_{\theta}}$, where $N_{\theta}$ denotes the number of actuated joints excluding the gripper and equals $3$ in the robotic arm configuration.

In the simplified model, the translational motion is determined by the total thrust $\tilde{f}_b$ generated by the aerial base, the gravitational acceleration $\bar{g}$, and the disturbance force $\mathbf{f}_d$ induced by the robotic arm.
The rotational motion is determined by the control torque $\boldsymbol{\tau}_b$ generated by the base, the gyroscopic effect, and the disturbance moment $\boldsymbol{\tau}_d$ induced by the robotic arm. 
The dynamic model of the simplified aerial manipulator can be expressed as:
\begin{equation}
\begin{cases}
\ddot{\mathbf{p}}_b = \tilde{f}_b\,\mathbf{R}_b \mathbf{e}_3 / m - \bar{g}\mathbf{e}_3 + \mathbf{f}_d / m, \\
\dot{\mathbf{R}}_b = \mathbf{R}_b \hat{\boldsymbol{\omega}}, \\
\mathbf{J}\dot{\boldsymbol{\omega}} = \boldsymbol{\tau}_b - \boldsymbol{\omega} \times (\mathbf{J}\boldsymbol{\omega}) + \boldsymbol{\tau}_d,
\end{cases}
\end{equation}
where $\mathbf{J}$ is the UAM inertia matrix and $\boldsymbol{\omega}$ is the body angular velocity.
$\mathbf{e}_3 = [0,\,0,\,1]^{\top}$ is the third column of the $3\times3$ identity matrix, and $\hat{(\cdot)}$ denotes the skew-symmetric operator for the cross product.
Given the base pose ${}^{w}\mathbf{T}_b$ and joint configuration $\boldsymbol{\theta}$, the pose of the $i$-th link is obtained by chaining the relative transforms along the kinematic chain:
\begin{equation}
{}^{w}\mathbf{T}_{i}(\boldsymbol{\theta})
=
{}^{w}\mathbf{T}_{b}\,
\prod_{k=1}^{i} {}^{k-1}\mathbf{T}_{k}(\theta_k),
\label{eq:link_pose_chain}
\end{equation}
where the product is ordered from $k=1$ to $k=i$, and the base frame
$b$ is used interchangeably with the frame $0$, i.e.,
${}^{w}\mathbf{T}_{b} \equiv {}^{w}\mathbf{T}_{0}$.

Inspired by quadrotor differential flatness, we use a flatness-based parameterization for the aerial base and optimize the manipulator joints jointly as additional trajectory variables. Accordingly, the flat output is defined as:
\begin{equation}
\mathbf{z} = (p_x, p_y, p_z, \psi, \boldsymbol{\theta})^{\top} \in \mathbb{R}^3 \times SO(2) \times \mathbb{R}^{N_{\theta}},
\end{equation}
which enables all system variables to be reconstructed from the derivatives of \(\mathbf{z}\).

\subsection{Collision Avoidance}
This subsection focuses on constructing safety constraints to ensure that the generated trajectory remains collision-free.
Since the UAM is required to directly interact with the environment, voxel-based or convex-hull representations would lead to a loss of geometric precision, which may easily result in collisions at the end-effector.
Therefore, we directly use the raw point cloud representing the environment and target objects as input for collision evaluation.

Based on the proposed MRC-SDF representation, safety constraints penalize configurations where any part of the UAM approaches surrounding obstacles. For each obstacle point $\mathbf{x}_j$ in the perceived point cloud, distances to all links are efficiently obtained by querying the link-wise SDFs (Sec.~\ref{sec:sdf}). We define the safety cost at time $t$ as:
\begin{equation}
J_c(t)
=
\sum_{j=0}^{N_p-1} \sum_{i=0}^{N_l-1}
\ell\!\left(d_{\mathrm{thr}} - {}^{i}d_j \right),
\label{eq:safety_cost}
\end{equation}
where $N_p$ denotes the number of obstacle points, 
$N_l$ denotes the number of link-wise SDFs used for collision checking, 
and $d_{\mathrm{thr}}$ is a predefined safety threshold. 
The penalty function $\ell(\cdot)$ is implemented as a smoothed $\ell_1$ loss:
\begin{equation}
\ell(x)=
\begin{cases}
0, & x < 0,\\[3pt]
x-\dfrac{\mu}{2}, & x > \mu,\\[8pt]
\left(\mu-\dfrac{x}{2}\right)\left(\dfrac{x}{\mu}\right)^3, & 0 \le x \le \mu,
\end{cases}
\label{eq:smoothed_l1}
\end{equation}
where $\mu>0$ is a smoothing parameter.

The gradient of the safety cost with respect to the state variables is computed via the chain rule as
\begin{equation}
\frac{\partial J_c}{\partial *}
=
\sum_{j=0}^{{N_p}-1} \sum_{i=0}^{{N_l}-1}
\frac{\partial \ell}{\partial {}^{i}d_j }
\frac{\partial {}^{i}d_j }{\partial {}^{i}\mathbf{x}_j}
\frac{\partial {}^{i}\mathbf{x}_j}{\partial *},
\label{eq:safety_gradient}
\end{equation}
where $*$ denotes the state variables $\{\mathbf{p}_b, \mathbf{R}_b, \boldsymbol{\theta}\}$. 
Here, $\partial {}^{i}d_j  / \partial {}^{i}\mathbf{x}_j$ is directly given by the spatial gradient of the corresponding link SDF, 
and $\partial {}^{i}\mathbf{x}_j / \partial *$ captures the dependence of the obstacle point coordinates on the UAM configuration through the forward kinematics.

This formulation enables efficient, fully differentiable collision-cost evaluation for whole-body trajectory optimization while preserving the geometric fidelity of the raw point cloud.
In the implementation, the gradients of the safety cost w.r.t. the state variables are given by
\begin{equation}
\begin{cases}
\displaystyle
\frac{\partial J_c}{\partial \mathbf{p}_b}
=
\sum_{j=0}^{N_p-1}
\sum_{i=0}^{N_l-1}
{}^{i}\alpha_j\,\mathbf{R}_i\,{}^{i}\mathbf{g}_j, \\[12pt]

\displaystyle
\frac{\partial J_c}{\partial \mathbf{q}}
=
-\sum_{j=0}^{N_p-1}
\sum_{i=0}^{N_l-1}
{}^{i}\alpha_j\,
{}^{b}\mathbf{g}_j^{\top}
\frac{\partial \mathbf{R}_b^{\top}}{\partial \mathbf{q}}
(\mathbf{x}_j - \mathbf{p}_b), \\[12pt]

\displaystyle
\frac{\partial J_c}{\partial \theta_k}
=
-\sum_{j=0}^{N_p-1}
\sum_{i=k+1}^{N_l-1}
{}^{i}\alpha_j\,
{}^{i}\mathbf{z}_k^{\top}
\Big(
{}^{i}\mathbf{g}_j \times ({}^{i}\mathbf{x}_j - {}^{i}\mathbf{p}_k)
\Big),
\end{cases}
\label{eq:safety_grad_compact}
\end{equation}
where ${}^{i}\mathbf{g}_j = \partial {}^{i}d_j  / \partial {}^{i}\mathbf{x}_j$ denotes the SDF gradient returned by the MRC-SDF query, 
${}^{i}\alpha_j = \partial \ell(d_{\mathrm{thr}} - {}^{i}d_j ) / \partial (d_{\mathrm{thr}} -{}^{i}d_j )$ is the derivative of the smooth hinge loss,
${}^{i}\mathbf{x}_j = \mathbf{R}_i^{\top}(\mathbf{x}_j-\mathbf{p}_i)$
is the obstacle point expressed in the link frame,
${}^{i}\mathbf{p}_k$ and ${}^{i}\mathbf{z}_k$ denote the position and rotation axis of the $k$-th joint in the link frame obtained from forward kinematics,
and $\mathbf{q} = [q_w,q_x,q_y,q_z]^{\top}$ denotes the unit quaternion parameterizing the base attitude.

\subsection{End-effector Insertion}
To facilitate stable straight-line insertion of the two-finger end-effector along the terminal segment until reaching the target pose, we design an end-effector insertion penalty. 
For a whole-body state with body pose $(\mathbf{p}_{b},\mathbf{q})$ and joint angles $\boldsymbol{\theta}$, the corresponding end-effector pose $(\mathbf{p}_e,\mathbf{q}_e)$ is computed via the forward-kinematics chain in Eq.~\ref{eq:link_pose_chain}. 
Given the desired end-effector goal pose $(\mathbf{p}_{e,g},\mathbf{q}_{e,g})$, where $\mathbf{p}_{e,g}$ and $\mathbf{q}_{e,g}$ denote the target position and orientation, respectively. We define the pose penalty as the sum of a position term and an orientation term, both using smoothed hinge penalties:
\begin{equation}
\begin{aligned}
J_{e} \;=\;& \rho_{P}\,\ell\!\left(\|\mathbf{p}_e-\mathbf{p}_{e,g}\| - d_{e,th}\right) \\
&+\; \rho_{R}\,\ell\!\left(d_R(\mathbf{R}_e,\mathbf{R}_{e,g}) - r_{e,th}\right),
\end{aligned}
\label{eq:end_pose_cost_k}
\end{equation}
where $\rho_{P}>0$ and $\rho_{R}>0$ are weighting coefficients, $d_{e,th}$ and $r_{e,th}$ are the position and rotation tolerances, respectively, and $\ell(\cdot)$ denotes the smoothed function defined in Eq.~\ref{eq:smoothed_l1}.
$\mathbf{R}_e$ and $\mathbf{R}_{e,g}$ are the rotation matrices corresponding to $\mathbf{q}_e$ and $\mathbf{q}_{e,g}$, respectively.

The end-effector insertion penalty $J_e$ is differentiable w.r.t.\ the whole-body state through the forward-kinematics chain in Eq.~\ref{eq:link_pose_chain}. 
Denoting $\mathbf{d_p}=\mathbf{p}_e-\mathbf{p}_{e,g}$ and $d_R=\frac{3-\mathrm{tr}(\mathbf{R}_e\mathbf{R}_{e,g}^{\top})}{2}$, the gradients used in the implementation are

\begin{equation}
\begin{cases}
\displaystyle
\frac{\partial J_e}{\partial \mathbf{p}_b}
=
\rho_P\,\alpha_P\,\mathbf{g}_p.
\\[10pt]

\displaystyle
\frac{\partial J_e}{\partial \mathbf{q}} = \rho_P\,\alpha_P\, \mathbf{J}^{p}_{\mathbf{q}}{}^{\top}\mathbf{g}_p \;+\; \rho_R\,\alpha_R\, \mathbf{g}_{\mathbf{q}},
\\[10pt]

\displaystyle
\frac{\partial J_e}{\partial \theta_k}
=
\rho_P\,\alpha_P\,
\mathbf{J}^{p}_{\theta_k}{}^{\top}\mathbf{g}_p
\;+\;
\rho_R\,\alpha_R\,
g_{\theta_k},
\end{cases}
\label{eq:Je_grad_compact}
\end{equation}
where $\mathbf{g}_p = \mathbf{d}_p / \lVert \mathbf{d}_p \rVert$, and $\alpha_{*}$ is the derivative of the smoothed hinge loss.
$\mathbf{J}^{p}_{\mathbf{q}}$, and $\mathbf{J}^{p}_{\theta_k}$ are the Jacobians of $\mathbf{p}_e$ w.r.t.\ $\mathbf{q}$, and $\theta_k$. 
For the rotation term, letting $\mathbf{G}_R=-\frac{1}{2}\mathbf{R}_{e,g}$, we use
\begin{equation}
\begin{aligned}
\mathbf{g}_{\mathbf{q}}
=
\Big[
&\mathrm{tr}\!\left(\mathbf{G}_R^{\top}\frac{\partial \mathbf{R}_e}{\partial q_w}\right),\;
\mathrm{tr}\!\left(\mathbf{G}_R^{\top}\frac{\partial \mathbf{R}_e}{\partial q_x}\right), \\
&\mathrm{tr}\!\left(\mathbf{G}_R^{\top}\frac{\partial \mathbf{R}_e}{\partial q_y}\right),\;
\mathrm{tr}\!\left(\mathbf{G}_R^{\top}\frac{\partial \mathbf{R}_e}{\partial q_z}\right)
\Big]^{\top}, \\
g_{\theta_k}
=&\;\mathrm{tr}\!\left(\mathbf{G}_R^{\top}[\mathbf{z}_k^{w}]_{\times}\mathbf{R}_e\right),
\end{aligned}
\label{eq:Je_rot_grad_terms}
\end{equation}
with $\frac{\partial \mathbf{R}_e}{\partial q_j}=\frac{\partial \mathbf{R}_b}{\partial q_j}\mathbf{R}_{e}^{b}$ and $\frac{\partial \mathbf{R}_e}{\partial \theta_k}=[\mathbf{z}_k^{w}]_{\times}\mathbf{R}_e$.

\subsection{Dynamic Feasibility}
To ensure that the generated trajectory can be physically executed by the aerial manipulator, 
it must satisfy the underlying dynamic constraints of the system. 
These constraints are enforced by penalizing violations in translational, rotational, and joint motion, 
as represented by the following unified cost term:
\begin{equation}
\mathcal{J}_*(t) = g\big(\|*(t)\|^2 - {*_{\max}}^2\big),
\end{equation}
where \(* \in \{v, a, \dot{\psi}, \dot{\theta}_k\}\) 
represents the translational velocity, acceleration, yaw rate, and the angular velocity of the \(k\)-th actuated joint, respectively, with \(k=1,\ldots,N_{\theta}\).

In addition, we further penalize the body rates around the \(x\)- and \(y\)-axes, 
denoted by \(\boldsymbol{\omega}_{xy} = [\omega_x, \omega_y]^T\), 
to ensure that the rotational motion remains within the feasible actuation range. 
The corresponding penalty function is defined as
\begin{equation}
\mathcal{J}_{\omega_{xy}} = 
\mathcal{L}_\mu \big(\|\boldsymbol{\omega}_{xy}\|^2 - \omega_{xy,\max}^2\big),
\end{equation}
where \(\omega_{xy,\max}\) denotes the maximum allowable body angular velocity 
along the \(xy\)-plane. 

Following~\cite{yuman24tro}, we adopt the Hopf fibration formulation to simplify computation and avoid singularities, allowing the body rates about the \(x\)- and \(y\)-axes to be computed directly from the time derivative of the body \(z\)-axis:
\begin{equation}
\|\boldsymbol{\omega}_{xy}\|^2 = \omega_x^2 + \omega_y^2 = \|\dot{\mathbf{z}}_b\|.
\end{equation}
The term \(\dot{\mathbf{z}}_b\) can be computed from the drone’s acceleration and jerk.

\subsection{Trajectory Representation and Optimization Problem Formulation}
We represent and optimize the motion plan using the MINCO trajectory class
\(\mathfrak{T}_{\text{MINCO}}^s\)~\cite{zhepei22tro}.
An \(s\)-order MINCO trajectory is expressed as a polynomial of degree \((2s-1)\).
It has been shown to yield the unique optimum of the minimum-control problem for an \(s\)-integrator chain, while remaining continuously differentiable up to order \(s-1\).
This representation admits a compact parametrization through intermediate waypoints and segment durations, denoted by \(\{\mathbf{q}, \mathbf{T}\}\).
Accordingly, trajectory generation is posed as the following optimization:
\begin{equation}
\min_{\mathbf{q}, \mathbf{T}}
\quad J_s + \rho T + \int_{0}^{T} J_G \, dt,
\end{equation}
where \(J_s\) is the smoothness objective derived directly from the polynomial trajectory, \(\rho\) is a time-regularization weight, \(T\) denotes the total time horizon, and \(J_G\) captures the costs arising from the UAV dynamic-feasibility, safety, and insertion constraints.

The detailed formulations of each term are as follows:
\begin{equation}   
\mathcal{J}_{\mathcal{G}} = 
\lambda_{\mathcal{G}}^{\top}
\big[ J_c,\, J_e,\, J_v,\, J_a,\, J_{\dot{\psi}},\, J_{\dot{\theta}},\, J_{\omega_{xy}}\big]^{\top},
\end{equation}
where $\lambda_{\mathcal{G}}$ is a predefined weight vector. $J_c$ denotes the collision-avoidance cost, and $J_e$ is the end-effector insertion cost. The terms $J_v$, $J_a$, $J_{\dot{\psi}}$, $J_{\dot{\theta}}$, and $J_{\omega_{xy}}$ penalize violations of the bounds on $v$, $a$, $\dot{\psi}$, $\dot{\theta}_k$, and the body angular velocity in the $x$--$y$ plane, respectively, where $\omega_{xy}=[\omega_x,\,\omega_y]^{\top}$.

We adopt $\mathfrak{T}_{\text{MINCO}}^{4}$ to parameterize the position trajectory $\mathbf{p}_b(t)=(p_x(t),p_y(t),p_z(t))^{\top}$, and $\mathfrak{T}_{\text{MINCO}}^{2}$ to parameterize the yaw trajectory $\psi(t)$ and the joint-angle trajectories $\boldsymbol{\theta}(t)$.
The initial guesses for $\psi(t)$ and $\boldsymbol{\theta}(t)$ are obtained by linear interpolation over time between their values at the initial and target configurations.

\begin{figure*}[!t]
	\begin{center}
		\includegraphics[width=2.0\columnwidth]{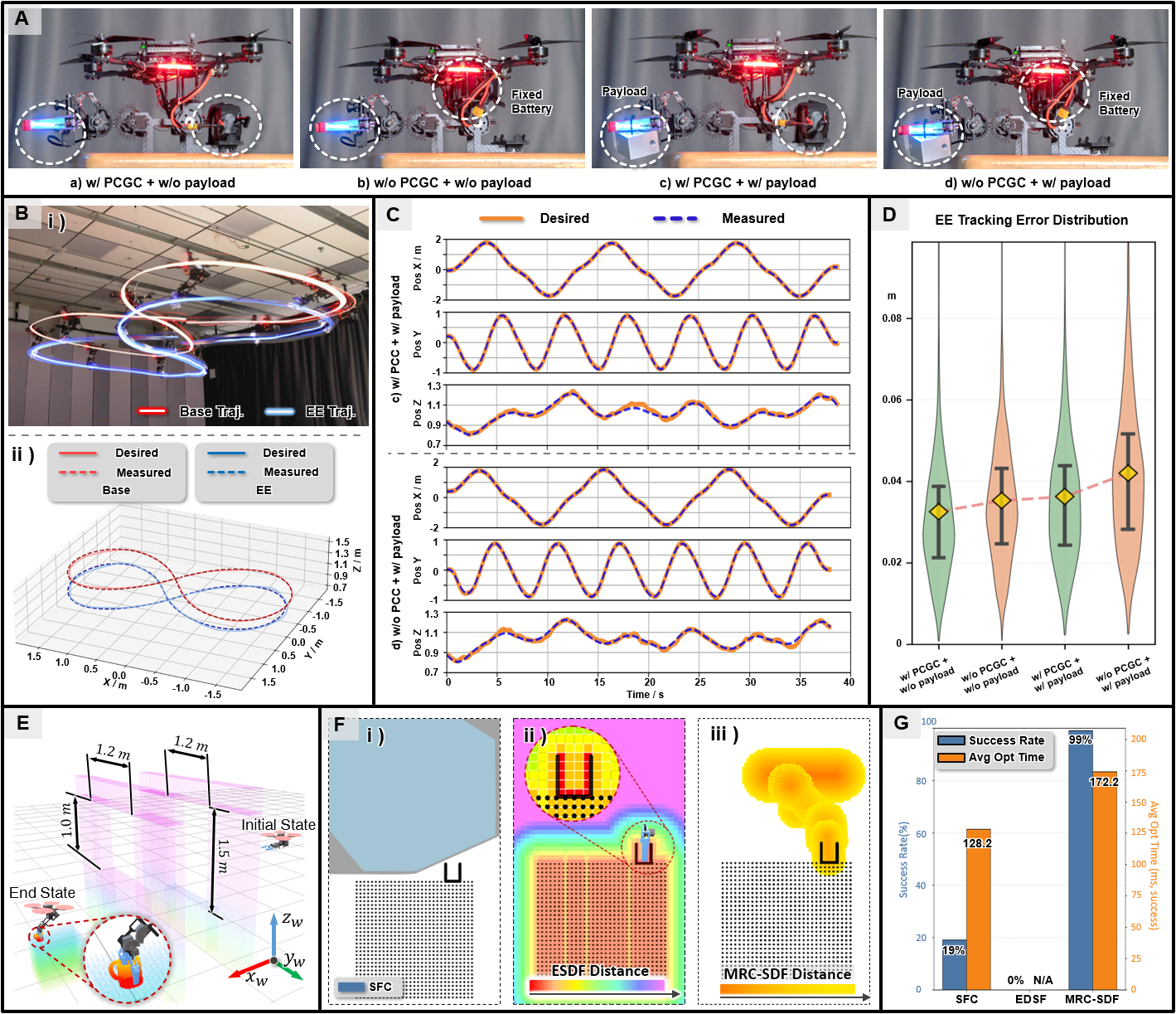}
	\end{center}
    \caption{
    \label{fig4} Trajectory tracking experiments. 
    (A) Experimental configurations. 
    (B) Long-exposure image of the figure-eight trajectory and the corresponding position tracking curves. 
    (C) Tracking curves with payload under configurations with and without PCGC. 
    (D) Distribution of three-dimensional absolute tracking errors over three trials of the figure-eight trajectory, with three loops per trial.
    (E) Benchmark comparison setup.
    (F) Comparison of different geometric representations: SFC, ESDF, and MRC-SDF.
    (G) Comparison of success rate and average optimization time for successful trials.
    }
	\vspace{-0.5cm}
\end{figure*}

\section{Experiments}
\subsection{Implementation Details}
In the real-world experiments, the proposed algorithm is deployed on the custom experimental platform QuadHand integrating an NVIDIA Jetson Orin NX as the onboard computer and an NxtPX4v2 flight controller. 
The root pitch link of the arm is actuated by a motor, while the remaining links, including the gripper, are driven by servos.
State estimation is performed using an EKF that fuses measurements from a Vicon motion capture system and an onboard IMU.
In the simulation experiments, the algorithm is executed on a desktop equipped with an Intel® Core™ Ultra 9 185H CPU.

\subsection{Figure-Eight Trajectory Tracking}
To assess the platform’s maneuverability, tracking performance, and the effectiveness of PCGC, the quadrotor is commanded to follow a figure-eight trajectory of $3.6 \times 2$~m while the arm joints execute sinusoidal motions within their respective limits, as shown in Fig.~\ref{fig4}. As illustrated in Fig.~\ref{fig4}A, four experimental configurations are considered, corresponding to combinations with or without PCGC and with or without an external payload. The payload consists of two iron blocks with a total mass of 150~g attached to the gripper. For each configuration, the platform conducts three flights along a 3.6~m $\times$ 2~m figure-eight trajectory. The trajectory overlay and a representative lap under the with-PCGC-with-payload setting are shown in Fig.~\ref{fig4}B. Fig.~\ref{fig4}C presents representative tracking results for the payload case, comparing with and without PCGC. Fig.~\ref{fig4}D shows the distribution of the three-dimensional absolute tracking errors under different configurations, indicating that the errors are generally on the order of 3--4~cm and that PCGC effectively improves tracking accuracy.

\subsection{Simulation and Benchmark Comparisons}

To validate the efficiency and accuracy of MRC-SDF in trajectory optimization, comparative simulation experiments were conducted on  grasping tasks with obstacles, using SFC\cite{deng25tro} and ESDF\cite{chengkai24icra} as baselines.
As shown in Fig.~\ref{fig4}E, two window-shaped obstacle frames were vertically placed at different heights in the environment.
The proposed system was required to start from the right side of the scene, pass through the windows, and fly toward the target position on the left to grasp the cup on the cube. The initial position of the UAM was randomly initialized within a $1 \times 2 \times 1\,\mathrm{m}^3$ box, while the joint configuration was randomly sampled within its executable range. So that the corresponding average required flight distance was approximately $4.5\,\mathrm{m}$. To ensure safe flight and grasping, the safe distance, defined as the minimum distance between the UAM surface and obstacles, was set to $0.03\,\mathrm{m}$.

\begin{figure*}[!t]
	\begin{center}
		\includegraphics[width=2.0\columnwidth]{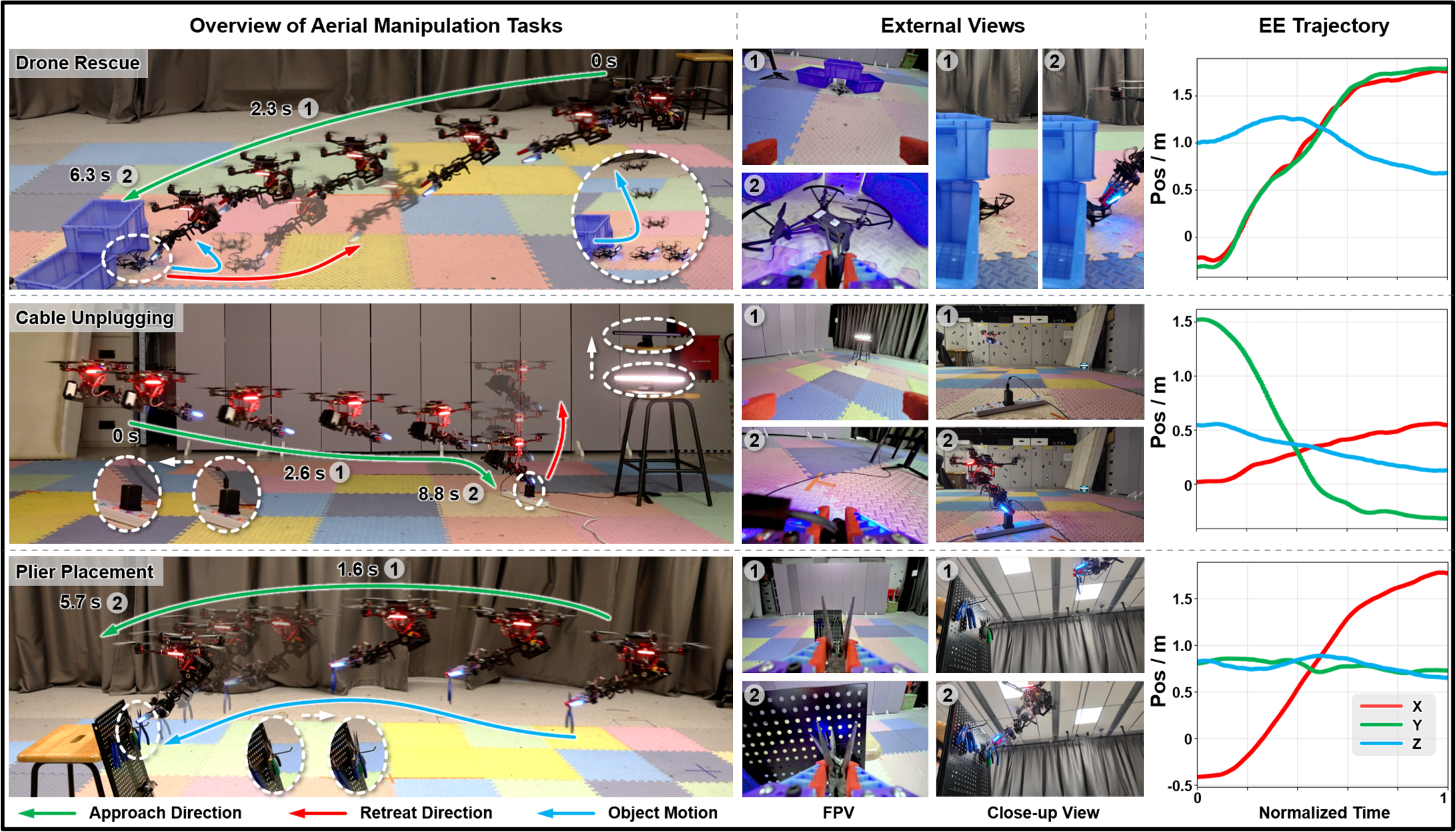}
	\end{center}
    \caption{
        \label{fig5}
        General mobile manipulation tasks: Drone Rescue, Cable Unplugging, and Plier Placement.
        Each row shows the motion trail, FPV images, external views, and EE position trajectories.
        Green, red, and blue arrows denote approach, retreat, and object motion, respectively.
    }
    
	\vspace{-0.5cm}
\end{figure*}

As shown in Fig.~\ref{fig4}F-i, in the SFC method, all environmental point clouds were considered to construct convex corridors for safety, which excluded the region around the target object from the feasible space. As a result, large penalties were imposed near the end of the trajectory, making it difficult to generate feasible grasping trajectories for close-range interaction.

For ESDF, the resolution was set to $0.02\,\mathrm{m}$ to balance geometric representation accuracy and onboard computational load. As shown in Fig.~\ref{fig4}F-ii, the gradient inside the cup was not resolved sufficiently, which prevented the optimizer from finding feasible trajectories that satisfied the safety constraint, resulting in the optimization failure.

In contrast, as shown in Fig.~\ref{fig4}F-iii, MRC-SDF provided a finer geometric representation when approaching obstacles while preserving the complete environmental point cloud. This avoided the loss of geometric details caused by voxelization and ensured the success of trajectory optimization.

Fig.~\ref{fig4}G shows the trajectory optimization statistics over 100 trials among the three methods. The success rate of the SFC corridor was 19\%, with an average optimization time of 128.24 ms, and the average SFC generation time was 29.27 ms. The success rate of ESDF was 0\%, and no feasible trajectory satisfying the safety constraints was found within the maximum optimization iterations. In addition, the average ESDF update time was 433.126 ms. The success rate of MRC-SDF reached 99\%  with an average optimization time of 172.22 ms under successful trials. These results indicate that MRC-SDF achieved the highest success rate while maintaining an acceptable optimization time.

\subsection{Real-world Implementation}
To validate the proposed system in real-world scenarios, we conduct five missions on the experimental platform. Drone Rescue, Cable Unplugging, and Plier Placement are presented in Fig.~\ref{fig5}, while gap-crossing grasp and narrow-gap crossing are presented in Fig.~\ref{fig:6}. In all missions, the terminal state is pre-specified, the initial state is estimated online, and the trajectory is computed onboard in real time.

\subsubsection{General Mobile Manipulation Tasks}
We conduct three mobile missions without considering the environment point cloud, as shown in Fig.~\ref{fig5}.
\begin{itemize}
    \item \textbf{Drone Rescue:} QuadHand approaches the grounded DJI Tello, grasps it, and drags it out from under the blue storage boxes; the Tello then takes off.
    \item \textbf{Cable Unplugging:} QuadHand navigates to the socket, grips the USB cable, and unplugs it, causing the cable-powered lamp to turn off.
    \item \textbf{Plier Placement:} QuadHand places a pre-grasped plier onto a hook on a perforated pegboard.
\end{itemize}

\subsubsection{Gap-crossing Related Experiments}
\begin{figure*}[!t]
    \centering
    \includegraphics[width=2.0\columnwidth]{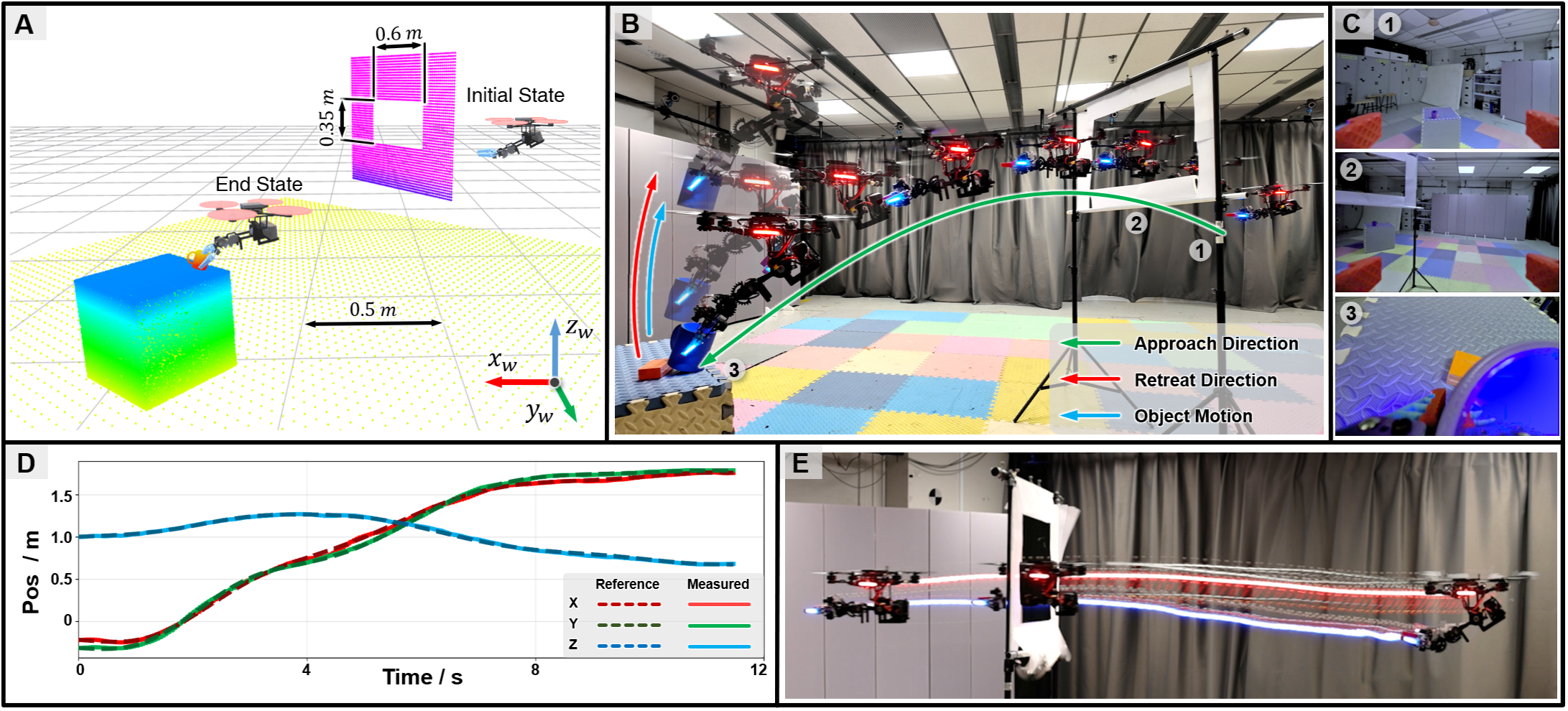}
    \caption{Gap-crossing experiments.
    (A--D) Gap-crossing grasping: setup, crossing-and-grasping snapshot, FPV image, and position tracking.
    (E) Long-exposure gap-crossing visualization.}
    \label{fig:6}
    \vspace{-0.5cm}
\end{figure*}

We conduct two gap-related experiments by considering the environment point cloud, as shown in Fig.~\ref{fig:6}. The narrow gap is sized at $0.6 \times 0.35~\mathrm{m}$.
\begin{itemize}
    \item \textbf{Narrow-gap crossing and cup grasping:} A cup (inner/outer diameter: 70/80~mm) is placed on a cube. QuadHand crosses the narrow gap, reaches above the cup, grasps it, and lifts it.
    \item \textbf{Narrow-gap crossing:} QuadHand is required to pass through the narrow gap from different initial states to the other side. We consecutively perform 7 trials, all of which successfully cross the gap.
\end{itemize}

\section{Conclusion and Future Work}
QuadHand integrates a compact quadrotor, 3-DoF arm, gripper, and battery-assisted passive CoG compensation module.
We also presented an MRC-SDF-based whole-body trajectory optimizer that preserves raw point-cloud detail for close-proximity planning.
Experiments demonstrate safe aerial manipulation in complex environments.
Current limitations include reliance on external motion capture and manually predefined terminal states and retreat directions.
Future work will study onboard state estimation and autonomous whole-body state generation and decision-making.

\bibliography{ICRA2022}
\end{document}